\def\year{2018}\relax
\documentclass[letterpaper]{article}
\usepackage{aaai18}
\usepackage{times}
\usepackage{lipsum}
\usepackage{helvet}
\usepackage{courier}
\usepackage{mathtools}
\usepackage{amsmath}
\usepackage{float}
\usepackage{graphicx}
\usepackage{algorithm}
\usepackage{algpseudocode}
\usepackage{amssymb}
\usepackage{array}
\usepackage{multirow}
\usepackage{booktabs}
\usepackage{siunitx}
\usepackage{tablefootnote}
\usepackage{amsthm}
\usepackage{url}
\usepackage[caption=false]{subfig}


\newcommand{\norm}[1]{\left\lVert#1\right\rVert}

\DeclarePairedDelimiter\ceil{\lceil}{\rceil}

\begin{document}

\title{Cross Temporal Recurrent Networks for Ranking Question Answer Pairs}
\author{Yi Tay\textsuperscript{$1$}, Luu Anh Tuan\textsuperscript{$2$} \and Siu Cheung Hui\textsuperscript{$3$}\\
\textsuperscript{$1,3$}\:Nanyang Technological University \\ School of Computer Science and Engineering, Singapore \\
\textsuperscript{$2$}\:Institute for Infocomm Research, Singapore \\
}

\maketitle
\begin{abstract}
Temporal gates play a significant role in modern recurrent-based neural encoders, enabling fine-grained control over recursive compositional operations over time. In recurrent models such as the long short-term memory (LSTM), temporal gates control the amount of information retained or discarded over time, not only playing an important role in influencing the learned representations but also serving as a protection against vanishing gradients. This paper explores the idea of learning temporal gates for sequence pairs (question and answer), jointly influencing the learned representations in a pairwise manner. In our approach, temporal gates are learned via 1D convolutional layers and then subsequently cross applied across question and answer for joint learning. Empirically, we show that this conceptually simple sharing of temporal gates can lead to competitive performance across multiple benchmarks. Intuitively, what our network achieves can be interpreted as learning representations of question and answer pairs that are aware of what each other is remembering or forgetting, i.e., pairwise temporal gating. Via extensive experiments, we show that our proposed model achieves state-of-the-art performance on two community-based QA datasets and competitive performance on one factoid-based QA dataset. 
\end{abstract}

\section{Introduction}

Learning-to-rank for QA (question answering) is a long standing problem in NLP and IR research which benefits a wide assortment of subtasks such as community-based question answering (CQA) and factoid based question answering. The problem is mainly concerned with computing relevance scores between questions and prospective answers and subsequently ranking them. Across the rich history of answer or document retrieval, statistical approaches based on feature engineering are commonly adopted. These models are largely based on complex lexical and syntactic features \cite{DBLP:conf/coling/WangM10a,DBLP:conf/acl/ZhouCZL11,DBLP:conf/sigir/WangMC09} and a learning-to-rank classifier such as Support Vector Machine (SVM) \cite{DBLP:conf/sigir/SeverynMTBR14,DBLP:conf/semeval/FiliceCMB16}.

 Today, we see a shift into neural question answering. Specifically, end-to-end deep neural networks are used for both automatically learning features and scoring of QA pairs. Popular neural encoders for neural question answering include long short-term memory (LSTM) networks \cite{hochreiter1997long} and convolutional neural networks (CNN). The key idea behind neural encoders is to learn to compose \cite{li2015visualizing}, i.e., compressing an entire sentence into a single feature vector. 

While it is possible to encode questions and answers independently, and later merge them with multi-layer perceptrons (MLP) \cite{DBLP:conf/sigir/SeverynM15}, tensor layers \cite{DBLP:conf/ijcai/QiuH15} or holographic layers \cite{DBLP:conf/sigir/tay2017}, it would be desirable for question and answer pairs to benefit from information available from their partner. There have been many models proposed for doing so which adopt techniques for jointly learning question and answer representations. Many of these recent techniques adopt soft-attention matching \cite{DBLP:conf/cikm/YangAGC16,DBLP:journals/corr/SantosTXZ16,DBLP:conf/aaai/ZhangLSW17} to learn attention weights that are jointly influenced by both question and answer. Subsequently, the joint attention weights are applied accordingly to learn a final representation of question and answer. Performance results have shown that incorporating the interactions between QA pairs can indeed improve the performance of QA systems. 

 Temporal gates form the cornerstone of modern recurrent neural encoders such as long short-term memory (LSTM) or gated recurrent units (GRU), serving as one of the key mitigation strategies against vanishing gradients. In these models, temporal gates control the inner recursive loop along with the amount of information being discarded and retained at each time step, allowing fine-grained control over the semantic compositionality of learned representations. Our work explores the idea of jointly learning temporal gates for sequence pairs, aiming to learn fine-grained representations of QA pairs which benefit from information pertaining to what each other is remembering or forgetting. 

The key idea here is as follows: \textit{By exploiting information about the question, can we learn an optimal way to semantically compose the answer?} (and vice versa). First, consider the following example in Table \ref{tab:example} which highlights the importance of semantic compositionality.

% Table generated by Excel2LaTeX from sheet 'Sheet5'
\begin{table}[htbp]
\small
  \centering
    \begin{tabular}{p{0.5cm}p{6cm}}
    \hline
    \hline
    Q: &  What \underline{deep learning framework} should I learn if I want to \textbf{get into} deep learning? I am a \textbf{beginner} without programming experience. Want to build cool apps.\\
    A: &  \underline{Tensorflow}. It is a pretty solid and low level deep learning framework for \textbf{advanced research}. \\
    \hline
    \hline
    \end{tabular}%
  
   \caption{Example of a question and answer pair. Ground truth is negative.}
   \label{tab:example}%
\end{table}%
First, it would be easy for many soft-attention and matching-based models to classify this question and answer pair with a high relevance score due to the underlined words (\underline{`deep learning'} and \underline{`tensorflow'}). However, the reason why this is a negative example is in the intricate details which can be effectively learned only via semantic compositionality (e.g., \textit{`without programming experience'}, \textit{`get into'}). On the other hand, by exploiting joint temporal gates, our method learns to compose the sentence given the information about its partner. For instance, without the knowledge of the answer which contains the phrase \textit{`advanced research'}, the question encoder will not know if it should retain the word \textit{`beginner'}. Hence, joint learning of temporal gates can help our model \textit{learn to compose}, by influencing what it remembers and forgets. As a result, this additional knowledge can allow the words in boldface (`\textit{beginner}',`\textit{advanced research}') to be strongly retained in the final representation. This is in similar spirit to neural attention. However, our approach jointly learns to compose instead of learning to attend. The difference is at the level which representations are influenced at. 

  \subsection{Our Contributions}
  The main contributions of this work are:

\begin{itemize}
\item We introduce a new method for using temporal gates to synchronously and jointly learn the interactions between text pairs. In the context of question answering, we learn which information to remember or discard in the answer while being aware of the context of the question. To the best of our knowledge, this is the first work that performs question answer matching at the temporal gate level. 
\item We propose a novel neural architecture for QA ranking. Our proposed Cross Temporal Recurrent Network (CTRN) model is largely inspired by the recently incepted Quasi Recurrent Neural Network (QRNN) \cite{DBLP:journals/corr/BradburyMXS16} and can be considered as a natural extension of QRNN to sequence pairs. Our model takes after QRNN in the sense that gates are first learned (via 1D convolutional layers) and then subsequently applied to temporally adjust the representations. Hence, the facilitation of information flow can be interpreted as \textit{joint pairwise gating}. 

\item Our proposed CTRN model achieves state-of-the-art performance on two community-based QA (CQA) datasets, namely the Yahoo Answers dataset and the QatarLiving dataset from SemEval 2016. Moreover, our model also achieves highly competitive performance on the TrecQA dataset for factoid based QA. Experimental results show that CTRN outperforms models that utilize attention-based matching while being significantly more efficient. Experimental results also confirm that CTRN improves the underlying QRNN model. 
\end{itemize}
\section{Related Work}
This section introduces prior work in the field of neural QA ranking. We also introduce the Quasi Recurrent Neural Network (QRNN) model, which lives at the heart of our proposed approach. 

\subsection{Neural Question Answer Ranking}
Convolutional neural network (CNN) \cite{DBLP:conf/nips/HuLLC14,DBLP:conf/sigir/SeverynM15} and recurrent models like the long short-term memory (LSTM) \cite{DBLP:conf/acl/WangN15} network are popular neural encoders for the QA ranking problem. \cite{DBLP:journals/corr/YuHBP14} proposed to use CNN for learning features and subsequently apply logistic regression for generating QA relevance scores. Subsequently, an end-to-end neural architecture based on CNN and bilinear matching was proposed in \cite{DBLP:conf/sigir/SeverynM15}. 

In many recent works, the key innovation in most models is the technique used to model interaction between question and answer pairs. The CNN model introduced in \cite{DBLP:conf/sigir/SeverynM15} uses a MLP to compose vectors of questions and answers. \cite{DBLP:conf/ijcai/QiuH15} adopted tensor layers for richer modeling capabilities. \cite{DBLP:conf/sigir/tay2017} proposed holographic memory layers. \cite{DBLP:conf/emnlp/HeGL15} proposed Multi-Perspective CNN which matches `multiple perspectives' based on variations of pooling and convolution schemes. Recent work has showed the effectiveness of learning QA embeddings in non-Euclidean spaces such as Hyperbolic space \cite{DBLP:journals/corr/TayLH17a}. Models based on soft-attention such as Attentive Pooling networks (AP-BiLSTM and AP-CNN) \cite{DBLP:journals/corr/SantosTXZ16}, AI-CNN \cite{DBLP:conf/aaai/ZhangLSW17} and aNMM (attention-based neural matching) \cite{DBLP:conf/cikm/YangAGC16} have also been proposed. These models learn weighted representations of QA pairs using similarity matrix based attentions.

\begin{figure*}[ht]
  \centering
    \includegraphics[width=0.9\linewidth]{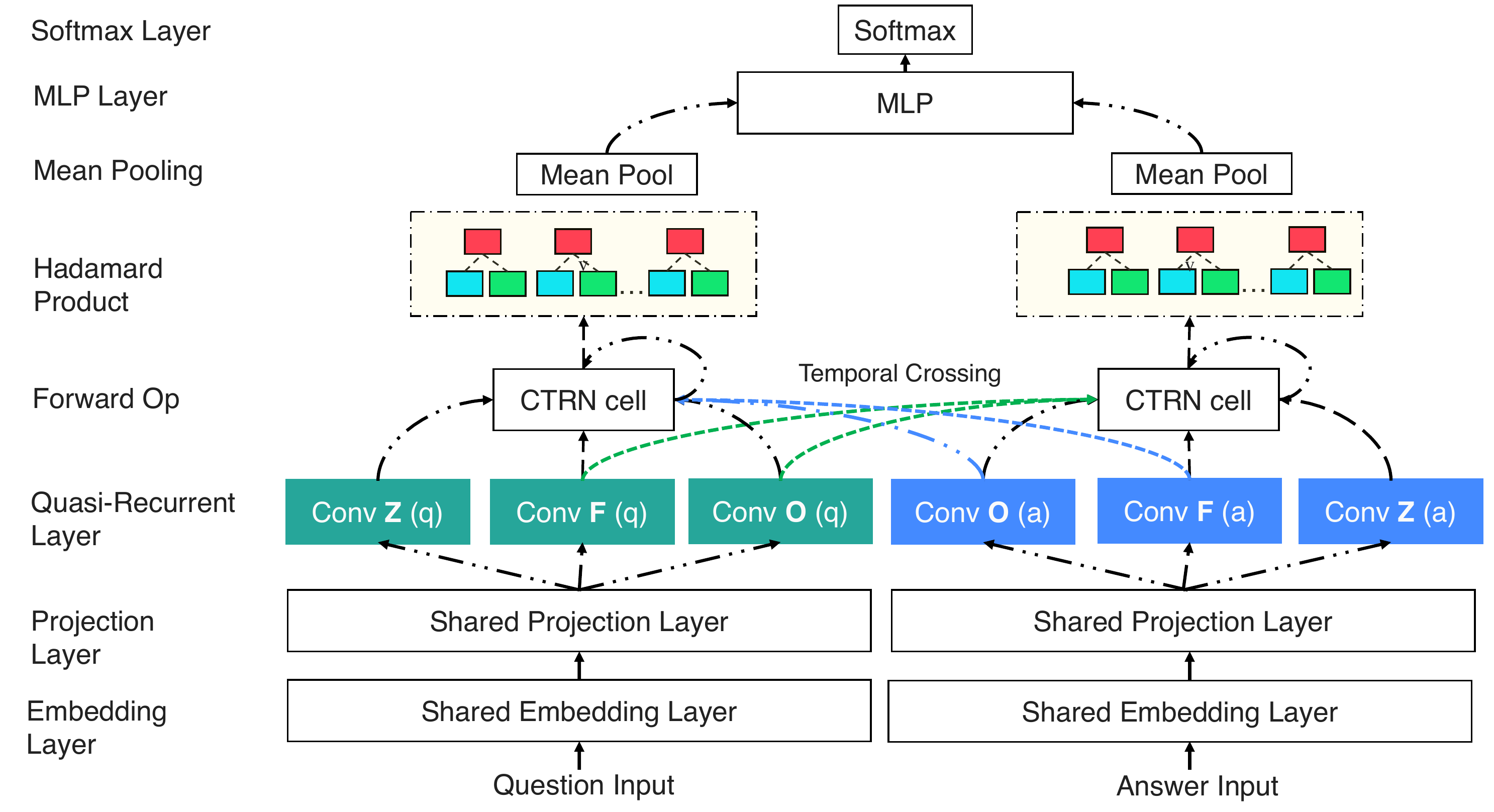}
    \caption{Diagram of our proposed CTRN architecture. Gates (denoted Conv F and Conv O) are prelearned via convolutions and each CTRN cell (for $q$ and $a$) incorporates the gates of their partner while learning representations. The base representations (in which gates are applied in CTRN) are denoted by Conv Z. Green denotes information flow from question and blue denotes information flow from answer.}
    \label{fig:CTRN_architecture}
\end{figure*}

\newcommand\numberthis{\addtocounter{equation}{1}\tag{\theequation}}
\subsection{Quasi Recurrent Neural Network (QRNN)}

In this section, we introduce Quasi Recurrent Neural Network \cite{DBLP:journals/corr/BradburyMXS16} and our key motivation of basing CTRN on QRNN. QRNN is a recurrent neural network that is actually a convolutional neural network in disguise. The key intuition with QRNN is that it first learns temporal gates via 1D convolutions and subsequently applies them sequentially. Contrastively, recurrent models learn these gates sequentially. Given an input of a sequence of $L$ vectors $w \in \mathbb{R}^{m}$ where $L$ is the maximum sequence length and $m$ is the dimension of the vectors, the QRNN model applies three 1D convolution operations as follows:
\begin{align*}% left aligned
\textbf{Z} &= tanh(\textbf{W}_z \ast \textbf{X}) \\ 
\textbf{F} &= \sigma (\textbf{W}_f \ast \textbf{X})  \numberthis \label{qrnn_eqn} \\
\textbf{O} &= \sigma (\textbf{W}_o \ast \textbf{X})
\end{align*}
where $\textbf{X}$ is the input sequence of $m$ dimensional vectors with sequence length $L$. $
\textbf{W}_x,\textbf{W}_f,\textbf{W}_o \in \mathbb{R}^{k \times d \times m}$ are parameters of QRNN and $\ast$ denotes a convolution across the temporal dimension. $k$ is the filter width and $d$ is the output dimension. Subsequently, the following equations describe the forward (recursive) operation of the QRNN cell:
\begin{align*}% left aligned
c_t &= f_t \odot c_{t-1} + (1-f_t) \odot z_t &\\  \notag
h_t &= o_t \odot c_t &
\end{align*}
where $c_t$ is the cell state and $h_t$ is the hidden state. $f_t,\: o_t$ are the forget and output gates respectively at time step $t$. $\sigma$ is the sigmoid activation which nonlinearly projects each element of its input to $[0,1]$. $\textbf{Z}$ can be regarded as the \textit{convolved} base representation similar to what a traditional CNN model learns. $\textbf{F}$ and $\textbf{O}$ are then applied recursively to temporally adjust and influence the semantic compositionality of $\textbf{Z}$. As such, this makes it \textit{`quasi-recurrent'}. The key difference between QRNN and recurrent models like LSTM is that gates are \textit{prelearned} via convolution while RNN models like LSTM learn their gates sequentially during the recursive forward operation. In short, the forward operation in QRNN is still sequentially applied but is comparatively much cheaper than traditional LSTM cells since gates are merely applied in the case of QRNN. As such, the parallelization of gate learning improves the speed of QRNN as compared to LSTM. In the original paper, QRNN achieved around 4 times less computational time as compared to LSTM models while achieving similar or better performance. For the sake of brevity, we refer interested readers to \cite{DBLP:journals/corr/BradburyMXS16} for more details. 

Inspired by the computational benefits of QRNN, we adopt it as our base model. Next, we also notice an attractive property of QRNN. In QRNN models, because gates are prelearned, it enables us to align temporal gates between two QRNNs easily. Conversely, considering the fact that questions and answers might not have similar sequence length, trying to sequentially align temporal gates in LSTM models can be extremely cumbersome and inefficient. More importantly, temporal gates of LSTM cells do not have global information, i.e., each step is only aware of all steps that precede it. On the other hand, temporal gates from QRNNs have global information about the entire sequence. 

\section{Our Proposed Approach}

In this section, we describe our novel deep learning model layer-by-layer. The overall architecture of our model is illustrated in Figure \ref{fig:CTRN_architecture}. 
For notational convenience, we denote the subscripts $q,a$ on whether a parameter belongs to question or answer respectively. 
\subsection{Embedding + Projection Layer}
Our model accepts two sequences of indices (question and answer inputs) which are passed through an embedding layer (shared between $q$ and $a$ inputs) and returns a sequence of $n$ dimensional vectors. In practice, we initialize and \textbf{fix} this layer with pretrained embeddings while connecting to a $n \times m$ projection layer. As such, the output of the embedding + projection layer is a $m$ dimensional vector. Note that this layer is shared between question and answer inputs. 

\subsection{Quasi-Recurrent Layer}
The input to the quasi-recurrent layer is a sequence of $L$ vectors $w \in \mathbb{R}^{m}$ where $L$ is the maximum sequence length. This layer applies three 1D convolution operations as described in Equation (\ref{qrnn_eqn}). Finally, the outputs of the quasi-recurrent layer are representations or matrices $\{\textbf{Z}_{s},\textbf{F}_{s},\textbf{O}_{s}\}$ where $s=\{q,a\}$. Note that up till now, this quasi-recurrent layer remains functionally identical to QRNN. 

\subsection{Lightweight Temporal Crossing (LTC)}
In this section, we introduce our novel lightweight temporal crossing (LTC) mechanism which lives at the heart of our CTRN model. Our approach extends upon the QRNN model, we leverage the fact that gates $\textbf{F}_x,\textbf{O}_x$ are learned non-sequentially. The key idea is to leverage the information in $\textbf{F}_q,\textbf{O}_q$ for $\textbf{Z}_a$ and vice versa. This information flow is denoted by the green and blue arrows in Figure \ref{fig:CTRN_architecture}. The outputs of this layer are similar to the LSTM model, i.e., they are a sequence of hidden states $\textbf{H} \in \mathbb{R}^{L \times d}$ where $L$ is the sequence length and $d$ is the number of filters. At this layer, there are two CTRN cells, namely CTRN-Q and CTRN-A, for question and answer representations respectively. 

\begin{figure}[ht]
  \centering
    \includegraphics[width=0.40\textwidth]{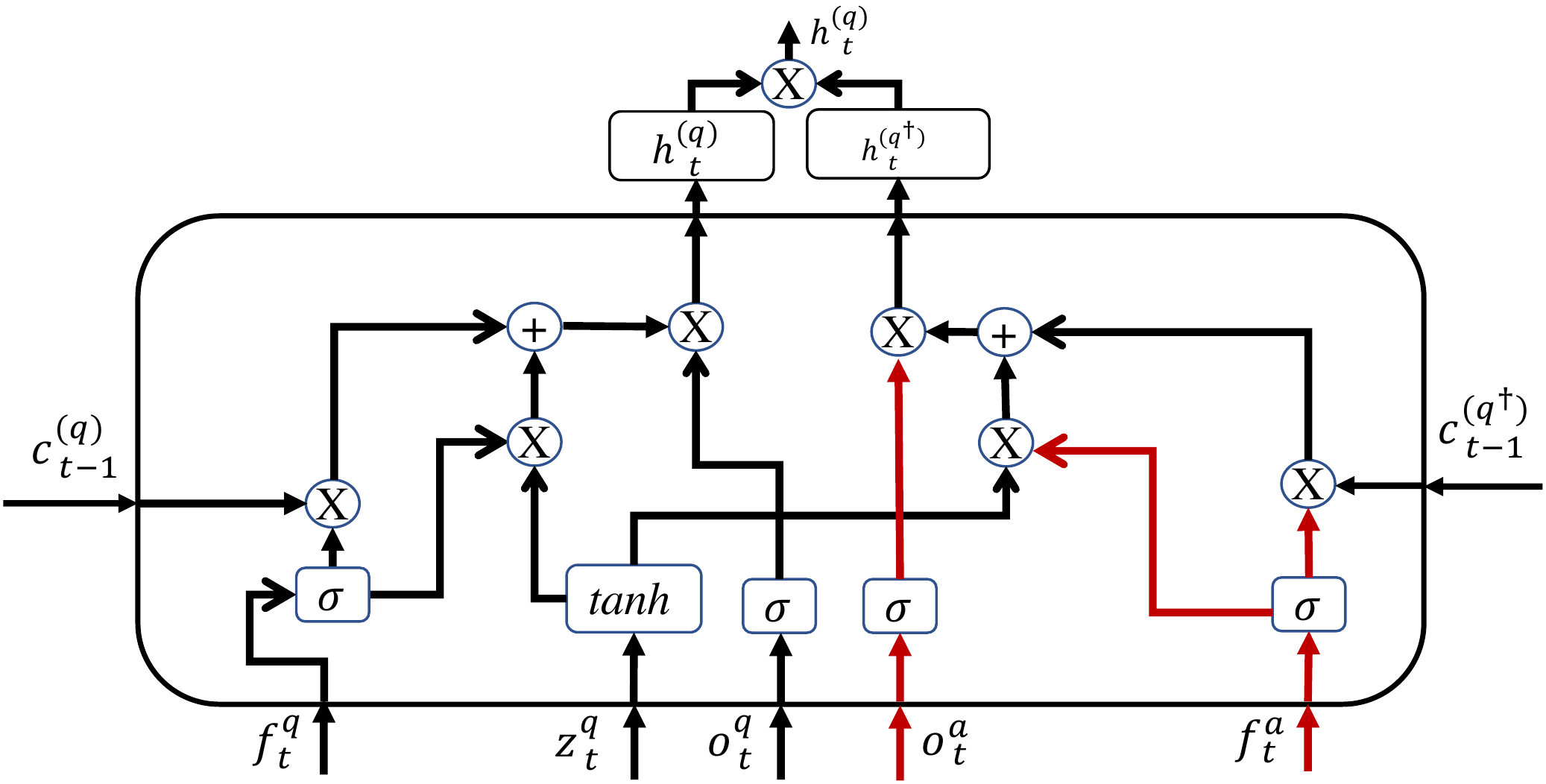}
    \caption{Diagram of a single CTRN-Q cell. Red lines are information flow from answer gates. $X$ denotes element-wise multiplication, $+$ denotes element-wise addition. $tanh$ is the hyperbolic tangent function and $\sigma$ is the sigmoid function.}
    \label{fig:CTRN_cell}
\end{figure}

In this section, we use CTRN-Q as an example but note that CTRN-A and CTRN-Q are functionally symmetrical. Figure \ref{fig:CTRN_cell} illustrates a single CTRN-Q cell. Each CTRN-Q cell contains two cell states denoted as $c^{(q)}_t$ and $c^{(q^{\dagger})}_t$ and two hidden states denoted as $h^{(q)}_t$ and $h^{(q^{\dagger})}_t$. As such, there are two learned representations in the CTRN-Q cell, denoted by $q$ and $q^{\dagger}$ respectively. The first representation is learned as per normal, i.e., applying $\textbf{F}_q, \: \textbf{O}_q$ on $\textbf{Z}_q$. The second representation is learned by applying partner gates $\textbf{F}_a, \: \textbf{O}_a$ on the question representation $\textbf{Z}_q$. The following equations depict the forward operation of the CTRN-Q cell. 
\begin{align*}% left aligned
c^{(q)}_t &= f^{q}_t \odot c^{(q)}_{t-1} + (1-f^{q}_t) \odot z^q_t & \\
h^{(q)}_t &= o^q_t \odot c^{(q)}_t &\\
c^{(q^{\dagger})}_t &= f^{a}_{t^{\ast}} \odot c^{(q^{\dagger})}_{t-1} + (1-f^{a}_{t^{\ast}}) \odot z^q_t&\\
h^{(q^{\dagger})}_t &= o^a_{t^{\ast}} \odot c^{(q^{\dagger})}_t &
\end{align*}
where $f^{n}_{t},o^{n}_{t},z^{n}_{t}$ denote the forget and output gates for text $n \in \{q,a\}$ at time step $t$. $t^{\ast}$ is an aligned time step between question and answer sequences as the sequence length of question and answer might be different. For simplicity, we consider $t^{\ast} = t \times \ceil{\frac{max(|q|,|a|)}{min(|q|,|a|)}}$. Similarly, the forward operation for CTRN-A is as follows:
\begin{align*}% left aligned
c^{(a)}_t &= f^{a}_t \odot c^{(a)}_{t-1} + (1-f^{a}_t) \odot z^a_t&\\
h^{(a)}_t &= o^a_t \odot c^{(a)}_t & \\
c^{(a^{\dagger})}_t  &= f^{q}_{t^{\ast}} \odot c^{(a^{\dagger})}_{t-1} + (1-f^{q}_{t^{\ast}}) \odot z^a_{t^{\ast}} &\\
h^{(a^{\dagger})}_t &= o^q_t \odot c^{(a^{\dagger})}_t  &
\end{align*}
Finally, to obtain a single representation for each question and answer. We simply apply the Hadamard product $\odot$ between hidden states of \textbf{each time step}, i.e., $h^{(q)}_{t} = h^{(q)}_{t} \odot h^{(q^{\dagger})}_{t}$ and $h^{(a)}_{t} = h^{(a)}_{t} \odot h^{(a^{\dagger})}_{t}$. This enables joint representations of temporal gates which form the crux of our LTC mechanism. 
\subsubsection{Why does this work?}
Notably, since gates are learned via parameterized convolutional layers, our learned gates ($\textbf{F}$ and $\textbf{O}$) not only contain `local' index-specific information but also `global' information of the entire text sequence. This is modeled by the parameters of the convolutional layers which produce $\textbf{F}$ and $\textbf{O}$. As such, it would suffice to compose them index-wise since the goal is to enable information flow between the temporal gates of question and answer. Our intuition here is to \textit{cross apply} question and answer gates to both question and answer representations so as to enable gradients flow across question and answer during back-propagation. Since the goal is to fuse and not to `match', we empirically found that soft-attention alignment of gates to yield no performance benefits over a simple index-wise alignment.

\subsubsection{Temporal Mean Pooling Layer}
The output of each CTRN cell is an array of hidden states $[h^{s}_1,h^{s}_2..h^{s}_L]$. In this layer, we apply temporal mean pooling for both CTRN-Q and CTRN-A. The operation of this layer is a simple element-wise average of all output hidden vectors.

\subsubsection{Dense Layers (MLP)}
The inputs of this layer are two vectors which are the final representations of question and answer respectively. In this layer, we concatenate the two vectors and pass them through a series of fully-connected dense layers (or MLP). Likewise, the number of layers is also a hyperparameter to be tuned. 

\subsubsection{Softmax Layer and Optimization}
The final output of the hidden layer is then passed through a 2-class softmax layer. The final score of each QA pair is described as follows:

\begin{equation}
s(q,a) = softmax(W_f \: x + b_f)
\end{equation}
where $x$ is the output of the last hidden layer and $W_f \in \mathbb{R}^{h \times 2}$ and $b_f \in \mathbb{R}^{2}$. $h$ is the size of the hidden layer. Our network minimizes the standard cross entropy loss as its training objective. The choice of a pointwise model is motivated by 1) ease of implementation and 2) previous work \cite{DBLP:conf/sigir/SeverynM15,DBLP:conf/sigir/tay2017,DBLP:conf/aaai/ZhangLSW17}. The loss function is defined as follows:

\begin{equation}
L = - \sum^{N}_{i=1} \: [ y_i \log s_i + (1-y_i)\log(1-s_i)] + \lambda\norm{\theta}^{2}_2
\end{equation}
\noindent where $s$ is the output of the softmax layer. $\theta$ contains all the parameters of the network and $ \lambda\norm{\theta}^{2}_2$ is the L2 regularization. The parameters of the network are updated using the Adam Optimizer \cite{DBLP:journals/corr/KingmaB14}. 

\section{Complexity Analysis}
In this section, we study the memory complexity of our model to further justify the \textit{lightweight} aspect in our LTC mechanism. First, our CTRN \textbf{does not} incur any parameter cost over the vanilla QRNN model ($3kdm$ for a single QRNN model). As such, the memory complexity and parameter size remain equal to QRNN. This is easy to see as there is no additional parameters added since our model, from a computational graph perspective, is simply adding connections between nodes. Next, we consider the runtime complexity of our model (forward pass). Let $d$ be the number of filters of the convolution layer and $L$ be the maximum sequence length. The computational complexity of a single QRNN cell is $\mathcal{O}(dL)$ excluding convolution operations used to generate $\{\textbf{F,\:O,\:Z}\}$. Though the number of operations is approximately doubled (due to cross applying gates), the complexity of a CTRN cell is still $\mathcal{O}(dL)$, i.e., our model still runs in linear time as compared to LSTM models with quadratic time complexity. Overall, our model, though seemingly more complicated, does not increase the parameter size and only incurs a slight increase in computational cost as compared to the already efficient QRNN model. We are also able to leverage the computational benefits of QRNN over the vanilla LSTM model. Table \ref{tab:complexity} shows a simple comparison of our proposed CTRN model against the standard LSTM and AP-BiLSTM models. We observe that QRNN and CTRN are much more parameter efficient as compared to recurrent models, only taking up $\approx 58\%$ the parameter size of the vanilla LSTM model and being $400\%$ smaller than AP-BiLSTM.

\begin{table}[htbp]
  \centering
  \small
    \begin{tabular}{|c|c|c|}
    \hline
    Model &  \# Mem Complexity & \# Params \\
    \hline
    LSTM & $4(md + d^2) + \textbf{2dh + h}$   & 1.79M\\
    AP-BiLSTM & $4(md + d^2) + \textbf{4d}^2$ & 5.86M      \\
    QRNN  &  $3 \: kdm + \textbf{2dh + h}$    & 1.05M \\
    CTRN  &  $3 \: kdm + \textbf{2dh + h}$    & 1.05M  \\
    \hline
    \end{tabular}%
    \caption{Memory complexity analysis with shared parameters for $q$ and $a$. $m$ is the size of the input embeddings. $d$ is the number of filters and the dimensionality of the LSTM model. $h$ is the size of the hidden layer. The complexity highighted in boldface is used to compose $q$ and $a$. \# Params gives an estimate with $d=512$, $m=300$, $h=128$ and $k=2$. Word embedding parameters are excluded from comparison. }
  \label{tab:complexity}%
\end{table}%

\section{Experiments}
To ascertain the effectiveness of our proposed approach, we conduct experiments on three popular benchmark datasets.

\subsection{Experimental Setup}
This section describes the datasets used, baselines compared and evaluation metrics.
\subsubsection{Datasets}
We select three popular benchmark datasets which are described as follows:
\begin{itemize}

\item \textbf{YahooQA} - Yahoo Answers is a CQA platform. This is a moderately large dataset containing $142,627$ QA pairs which are obtained from the CQA platform. More specifically, preprocessing and testing splits\footnote{Splits be obtained at \url{https://github.com/vanzytay/YahooQA_Splits}.} are obtained from \cite{DBLP:conf/sigir/tay2017}. In their setting, questions and answers that are not in the range of $5-50$ tokens are filtered. Additionally, $4$ negative samples are generated for each question by sampling from the top $1000$ hits using \textit{Lucene} search.

\item \textbf{QatarLiving} - This is another CQA dataset which was obtained from the popular SemEval-2016 Task 3 Subtask A (CQA). This is a real world dataset obtained from Qatar Living Forums. In this dataset, there are ten answers per thread (question) which are marked as `Good', `Potentially Useful' or `Bad'. Following \cite{DBLP:conf/aaai/ZhangLSW17}, we treat `Good' as positive and anything else as negative labels. 
\item \textbf{TrecQA} - This is a popular QA ranking benchmark obtained from the TREC QA Tracks 8-13. QA pairs are generally short and factoid-based consisting trivia like questions. In this dataset, there are \textbf{two} training sets, namely TRAIN and TRAIN-ALL. TRAIN consists of QA pairs that have been manually judged and annotated. TRAIN-ALL is an automatically judged dataset of QA pairs and contains a larger number of QA pairs. TRAIN-ALL, being a larger dataset, also contains more noise. Nevertheless, both datasets enable the comparison of all models with respect to the availability and volume of training samples.

\end{itemize}
The statistics of all datasets, i.e., training sets, development sets and testing sets, are given in Table \ref{tab:dataset}. 

\begin{table}[H]
  \centering
  \small

    \begin{tabular}{|c||cc||cc|}
    \hline

    & \multicolumn{2}{c}{CQA} & \multicolumn{2}{c|}{TrecQA} \\
    \hline
              & YahooQA & QL   & TRAIN & TRAIN-ALL\\
          \hline
    Train Qns &   50.1K         &  4.8K  &   94 &1229\\
    Dev Qns & 6.2K      &224    &  82 & 82 \\
    Test Qns & 6.2K    &  327   & 100 & 100\\
    \hline
    Train Pairs & 253K  &36K    & 4.7K & 53K\\
    Dev Pairs &31.7K    &2.4K  &   1.1K & 1.1K\\
    Test Pairs & 31.7K  & 3.2K    & 1.5K & 1.5K\\
   
    \hline
    \end{tabular}%
    \caption{Statistics of datasets. QL denotes the QatarLiving dataset. TRAIN and TRAIN-ALL are two settings of the TrecQA dataset. }
  \label{tab:dataset}%
\end{table}%

\subsubsection{Evaluation Metrics}
For each dataset, we adopt the evaluation metrics used in prior work. For YahooQA, we follow \cite{DBLP:conf/sigir/tay2017} that uses P@1 (Precision@1) and MRR (Mean Reciprocal Rank). For QatarLiving, we follow \cite{DBLP:conf/aaai/ZhangLSW17} and evaluate on P@1 and MAP (Mean Average Precision). For TrecQA, we follow the experiment procedure in \cite{DBLP:conf/sigir/SeverynM15} using the official evaluation metrics of MAP and MRR. Since the evaluation metrics are commonplace in ranking tasks, we omit any further details for the sake of brevity.

\subsubsection{Implementation Details and Baselines}
For our CTRN model, we tune the output dimension (number of filters) within $[128,1024]$ in multiples of $128$. A single layered CTRN and QRNN is used. The number of dense (MLP) layers is tuned from $[1,3]$ and learning rate tuned amongst $\{10^{-3}, 10^{-4}, 10^{-5}\}$. Batch size is tuned amongst $\{64,128,256,512\}$. Dropout is set to $0.5$ and L2 regularization is set to $4 \times 10^{-6}$. Word embedding matrices are all non-trainable and are learned by the projection layer instead. For the three datasets, we adopt dataset-specific baselines largely based on prior published works. 

\begin{itemize}
\item \textbf{YahooQA} - We compare against multiple state-of-the-art models. Specifically, we compare our model with the vanilla LSTM, vanilla CNN, CNTN, NTN-LSTM and HD-LSTM. Since we use the same testing splits, we report the results directly from \cite{DBLP:conf/sigir/tay2017}. Additionally, we include additional baselines such as AP-BiLSTM and AP-CNN \cite{DBLP:journals/corr/SantosTXZ16} which serve as a representative for soft-attention alignment based models. Cosine similarity with pairwise ranking is used as the metric for AP-BiLSTM and AP-CNN following the original implementation. Models superscripted with $^\dagger$ are implemented by us. We initialize our model with pretrained GloVE embeddings \cite{DBLP:conf/emnlp/PenningtonSM14} of $d=300$.
\item \textbf{QatarLiving} - The key competitors of this dataset are the CNN-based ARC-I/II architecture by Hu et al. \cite{DBLP:conf/nips/HuLLC14}, the Attentive Pooling CNN \cite{DBLP:journals/corr/SantosTXZ16}, Kelp \cite{DBLP:conf/semeval/FiliceCMB16} a feature engineering based SVM method, ConvKN \cite{DBLP:conf/semeval/Barron-CedenoMJ16} a combination of convolutional tree kernels with CNN and finally  AI-CNN (Attentive Interactive CNN) \cite{DBLP:conf/aaai/ZhangLSW17}, a tensor-based attentive pooling neural model. We initialize with pretrained GloVE embeddings of $d=200$ trained using the domain-specific unannotated corpus provided by the task. 

\item \textbf{TrecQA} - We compare against published works which include both traditional models and neural models. Moreover, we compare with models reported in \cite{DBLP:conf/sigir/tay2017} on TRAIN and TRAIN-ALL datasets to observe the effect of different dataset sizes. The evaluation procedure follows \cite{DBLP:conf/sigir/SeverynM15} closely. We initialize the embedding layers with the same pretrained word embeddings of $d=50$ as \cite{DBLP:conf/sigir/SeverynM15} for fair comparisons against competitor approaches. These embeddings are trained with the Skip-gram model using the Wikipedia and AQUAINT corpus. \textit{Four} word overlap features are also concatenated before the dense layers following \cite{DBLP:conf/sigir/SeverynM15}. We train our model for $25$ epochs for TRAIN and $5$ epochs for TRAIN-ALL and report the test score from the best performing model on the development set. Hyperparameters are also tuned on the development set. Early stopping is adopted and training is terminated if the validation performance doesn't improve after $5$ epochs. 

\end{itemize}

\subsection{Experimental Results}
In this section, we report some observations pertaining to our empirical results. 

\subsubsection{Experimental Results on YahooQA}

\begin{table}[ht]
\small
  \centering

    \begin{tabular}{|c||cc|}
    
    \hline
          
         \textbf{Model} & \textbf{P@1}   & \textbf{MRR} \\
          \hline
    Random Guess & 0.200 & 0.457 \\
    BM-25 & 0.225 & 0.493 \\
    CNN$^\phi$  & 0.413 & 0.632 \\
    CNTN$^\phi$  & 0.465 & 0.632  \\
    LSTM$^\phi$ & 0.465 & 0.669  \\
    NTN-LSTM$^\phi$ & 0.545 & 0.731 \\
    HD-LSTM$^\phi$ & 0.557 & 0.735 \\
    AP-CNN$^\dagger$ &0.560 & 0.726\\
    AP-BiLSTM$^\dagger$ & 0.568 & 0.731 \\
     
    QRNN$^\dagger$ & \underline{0.573} & \underline{0.736}\\
    \hline
    CTRN (This paper)  & \textbf{0.601}& \textbf{0.755}\\

    \hline
    
    \end{tabular}%
     \caption{Experimental results on YahooQA. Models are ranked by P@1. Models marked with $\phi$ are reported directly from \cite{DBLP:conf/sigir/tay2017} while $\dagger$ denotes our own implementation. Best result is in boldface and second best is underlined. }
  \label{tab:yahooQAresults}%
\end{table}%

Table \ref{tab:yahooQAresults} reports the experimental results on the YahooQA dataset. Firstly, we observe that
our proposed CTRN achieves state-of-the-art performance on this dataset. Notably, we outperform HD-LSTM \cite{DBLP:conf/sigir/tay2017} by
$4\%$ in terms of P@1 and $2\%$ in terms of MRR. CTRN also outperforms attention based models such as AP-BiLSTM and AP-CNN \cite{DBLP:journals/corr/SantosTXZ16} by a considerable margin, i.e., of about $2\%-3\%$. At this junction, we make several observations about our proposed CTRN model. Firstly, this shows that our LTC mechanism is more effective than soft-attention matching on this dataset. Secondly, the merits of this mechanism can be further observed by the performance difference in the QRNN and CTRN. Our proposed CTRN comfortably outperforms QRNN by $2\%-3\%$ in terms of P@1 and MRR. Surprisingly, we see that a simple baseline QRNN performs quite well on this dataset which outperforms other complex models such as NTN-LSTM and HD-LSTM \cite{DBLP:conf/sigir/tay2017}.

\subsubsection{Experimental Results on QatarLiving}
Table \ref{tab:qatar} reports our experimental results on the QatarLiving dataset. Our CTRN model outperforms AI-CNN\footnote{For fair comparison, we compare against the reported results of AI-CNN that does not use handcrafted features.} by $2.5\%$ in terms of P@1 while maintaining similar performance on MRR. The performance of the CTRN model also outperforms the baseline QRNN by $3\%$ on P@1. Similar to the Yahoo QA dataset, we also found that the baseline QRNN performed surprisingly well, i.e., outperforming ConvKN and other CNN based models such as ARC-I and ARC-II. Overall, our proposed approach achieves very competitive results on this dataset. 
\begin{table}[htbp]
  \centering
\small
    \begin{tabular}{|c||c|c|}
    
    \hline
    \textbf{Model} & \textbf{P@1}   & \textbf{MAP} \\
    \hline
    ARC-I CNN & 0.741 & 0.771 \\
    ARC-II CNN & 0.753 & 0.780 \\
    AP    & 0.755 & 0.771 \\
    Kelp  & 0.751 & 0.792 \\
    ConvKN & 0.755 & 0.777 \\
    QRNN & 0.758& 0.783 \\
    AI-CNN & \underline{0.763} & \underline{0.791} \\
    \hline

    CTRN (This paper) & \textbf{0.788} & \textbf{0.794} \\
    \hline
    
    \end{tabular}%
     \caption{Experimental results on the QatarLiving dataset. Best result is in boldface and second best is underlined. CTRN outperforms the complex AI-CNN model. }
  \label{tab:qatar}%
\end{table}%

\subsubsection{Results on TrecQA}
\begin{table}[H]
\small
  \centering
    \begin{tabular}{|c||cc|cc|}

       \hline
          & \multicolumn{2}{c}{\textbf{TRAIN}} & \multicolumn{2}{c|}{\textbf{TRAIN-ALL}} \\
          \hline
    \textbf{Model} & \textbf{MAP} & \textbf{MRR} & \textbf{MAP} & \textbf{MRR} \\
    \hline

 CNN + LR & 0.7058& 0.7846&0.7113 & 0.7846 \\
    CNN   &  0.7000 &0.7469 &0.7216 &0.7899  \\
    CNTN &  0.7045  &0.7562 &0.7278 & 0.7831   \\
      LSTM & 0.7007 &0.7777&  0.7350  &0.8064\\
    MV-LSTM &    0.7077 & 0.7821 & 0.7327 & 0.7940 \\

    NTN-LSTM &0.7225 &0.7904 & 0.7364& 0.8009  \\
    HD-LSTM &\underline{0.7520} &\underline{0.8146} & 0.7499& 0.8153 \\
    QRNN  & 0.7000   & 0.7859 & \underline{0.7609} & \underline{0.8227} \\
    \hline
    CTRN (This paper) & \textbf{0.7582} & \textbf{0.8233} & \textbf{0.7712} & \textbf{0.8384} \\
    \hline
    
    \end{tabular}%
      \caption{Comparisons of various neural baselines on the TREC QA task on two dataset settings TRAIN and TRAIN-ALL. Competitors (except QRNN and CTRN) are reported from \cite{DBLP:conf/sigir/tay2017}. Best result is in boldface and second best is underlined.}
  \label{tab:results2}%
\end{table}%
\begin{table}[H]
   \centering
   \small
     \begin{tabular}{|l||cc|}

     \hline
     \multicolumn{1}{|c}{\textbf{Model}} & \textbf{MAP} & \textbf{MRR} \\
   
     \hline
     Wang et al. (2007) & 0.6029 & 0.6852 \\
     Heilman and Smith (2010)  & 0.6091 & 0.6917 \\
     Wang and Manning (2010)  & 0.5951 & 0.6951 \\
     Yao (2013)  & 0.6307 & 0.7477 \\
     Wang et al. (2015) - BM25 & 0.6370 &  0.7076 \\
     S \& M (2013)  & 0.6781 & 0.7358 \\
     Yih et al. (2013)  & 0.7092 & 0.7700 \\
     Yu et al. (2014) - CNN+LR  & 0.7113 & 0.7846 \\
     Wang et al. (2015) - biLSTM & 0.7143 & 0.7913 \\
     S \& M. (2015) - CNN  & 0.7459 & 0.8078 \\
     Yang et al. (2016) - aNMM & 0.7495 & 0.8109 \\
     Tay et al. (2017) - HD-LSTM & 0.7499 & 0.8153 \\
     Bradbury et al. (2016) - QRNN+MLP$^\dagger$&  0.7609& 0.8227 \\
     He and Lin (2016) - MP-CNN & \underline{0.7620}& \underline{0.8300}\\

     \hline
     CTRN$^\dagger$ (TRAIN) & 0.7582 & 0.8233 \\
     CTRN$^\dagger$ (TRAIN-ALL) & \textbf{0.7712} & \textbf{0.8384} \\
     \hline
     \end{tabular}%
     \caption{Performance comparisons of all published works on TREC QA dataset. $S \& M$ is short for Severyn and Moschitti. Best result is in boldface and second best is underlined. The MP-CNN result is reported from \cite{DBLP:conf/cikm/RaoHL16}, using the pointwise model for fair comparison. $\dagger$ denotes results reported by this work.}
   \label{tab:results_all}%
 \end{table}%{}

 Table \ref{tab:results2} reports the results on TRAIN and TRAIN-ALL settings of the TrecQA task. CTRN achieves the top results comparing to the multitude of neural baselines. Notably, the performance of CTRN is about $1\%-5\%$ better than QRNN. QRNN performs very competitively on the TRAIN-ALL setting but fails in comparison for the TRAIN setting. This might be because QRNN, with three 1D convolutional layers, might overfit on the smaller dataset. However, CTRN performs well on smaller TRAIN as well which hints at possibly some regularizing effect of the LTC mechanism. The performance of the vanilla QRNN model on the TRAIN-ALL setting is also surprisingly competitive, outperforming more complex models such as HD-LSTM and NTN-LSTM. 

Table \ref{tab:results_all} reports the results of the CTRN model against other published competitors. We can see that CTRN outperforms many complex neural architectures such as the aNMM model \cite{DBLP:conf/cikm/YangAGC16}, HD-LSTM \cite{DBLP:conf/sigir/tay2017} and MP-CNN model \cite{DBLP:conf/emnlp/HeGL15,DBLP:conf/cikm/RaoHL16}.

\subsection{Runtime Comparison}

\begin{figure}[ht]
  \centering
    \includegraphics[width=0.86\linewidth]{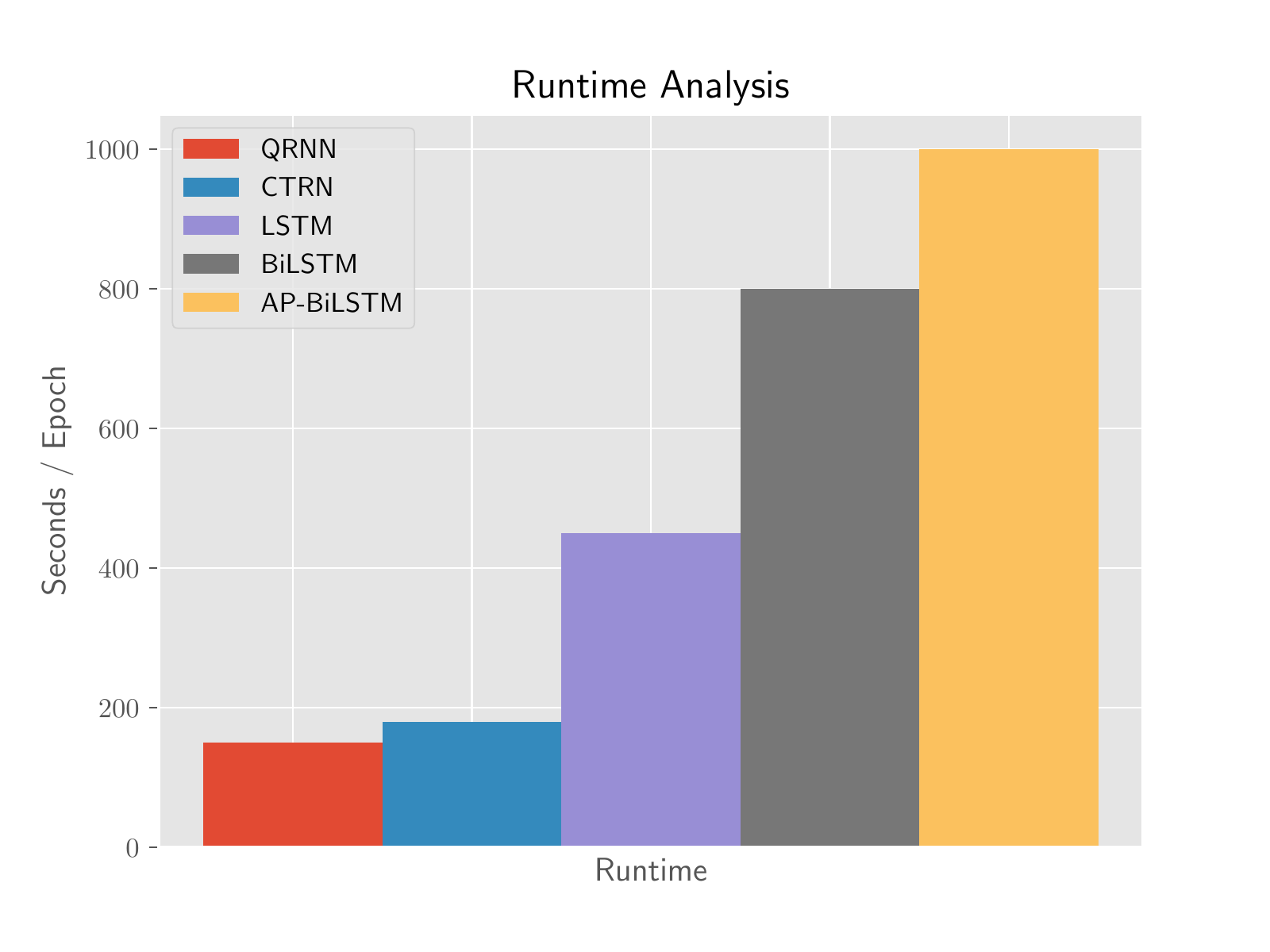}
    \caption{Comparisons of runtime of all recurrent models on the TRAIN-ALL dataset with $d=800$. }
    \label{fig:runtime}
\end{figure}

Figure \ref{fig:runtime} shows the runtime comparison for recurrent models on the TRAIN-ALL dataset. We observe that CTRN is a very scalable and efficient model. Notably, our CTRN model benefits from the training speed brought from the QRNN model, which is clearly significantly faster than LSTM models. Moreover, we also show that CTRN does not significantly increase the runtime of the base QRNN, only incurring an additional $\approx 10s$ per epoch. Moreover, we achieve $4$ times faster runtime compared to vanilla LSTM models and $8$ times faster than AP-BiLSTM.

\section{Conclusion}
We introduced a novel method for jointly learning to compose QA pairs. This is achieved by aligning temporal gates. We show that our lightweight temporal crossing (LTC) mechanism is an effective method of modeling interactions between QA pairs without incurring any parameter cost. Our CTRN model performs competitively on two CQA benchmarks and one factoid QA benchmark while being much faster than LSTM and AP-BiLSTM models.

\section{Acknowledgements}
The authors thank anonymous reviewers for their hardwork and feedback.

% \bibliography{references} 

\bibliography{references}
\bibliographystyle{aaai}
\end{document}